\documentclass[10pt,twocolumn,letterpaper]{article}

\usepackage{cvpr}              

\usepackage{multirow}

\definecolor{cvprblue}{rgb}{0.21,0.49,0.74}
\usepackage[pagebackref,breaklinks,colorlinks,allcolors=cvprblue]{hyperref}

\def\paperID{24} 
\def\confName{CVPR}
\def\confYear{2025}

\title{Multimodal Rationales for Explainable Visual Question Answering}

\author{Kun Li$^1$, George Vosselman$^1$, Michael Ying Yang$^2$\\
\textsuperscript{1}University of Twente, \textsuperscript{2}University of Bath\\
{\tt\small \textsuperscript{1}\{k.li, george.vosselman\}@utwente.nl, \textsuperscript{2}myy35@bath.ac.uk}}

\begin{document}
\maketitle
\begin{abstract}
Visual Question Answering (VQA) is a challenging task of predicting the answer to a question about the content of an image. 
Prior works directly evaluate the answering models by simply calculating the accuracy of predicted answers. 
However, the inner reasoning behind the predictions is disregarded in such a ``black box" system, and we cannot ascertain the trustworthiness of the predictions.
Even more concerning, in some cases, these models predict correct answers despite focusing on irrelevant visual regions or textual tokens.
To develop an explainable and trustworthy answering system, we propose a novel model termed \textbf{MRVQA} (\textbf{M}ultimodal \textbf{R}ationales for \textbf{VQA}), which provides visual and textual rationales to support its predicted answers.
To measure the quality of generated rationales, a new metric \textbf{vtS} (visual-textual Similarity) score is introduced from both visual and textual perspectives. 
Considering the extra annotations distinct from standard VQA, MRVQA is trained and evaluated using samples synthesized from some existing datasets. 
Extensive experiments across three EVQA datasets demonstrate that MRVQA achieves new state-of-the-art results through additional rationale generation, enhancing the trustworthiness of the explainable VQA model.
The code and the synthesized dataset are released under \url{https://github.com/lik1996/MRVQA2025}.
\end{abstract}    
\section{Introduction}
\label{sec:intro}
Visual Question Answering (VQA) takes as input an image and a natural language query and predicts the corresponding answer based on the understanding of the provided vision-language content.
It is more challenging than other fundamental computer vision tasks (\emph{e.g.,}~object detection \cite{liu2023grounding-dino}), since it requires fusion and reasoning cross different modalities.
With the rapid developments of deep neural networks, recent approaches \cite{ anderson2018bottom, dua2021beyondvqa,li2022blip, mao2022positionaltmm} achieve promising performance on VQA.
Normally, the answer accuracy \cite{antol2015vqa} is employed in most VQA systems to measure the correctness of the predictions.
However, users cannot know how reliable the predicted answers are, purely based on the accuracy numbers.
In contrast, explainable VQA (EVQA) tries to link the predicted answer to the given image-question pair by providing explanations in different formats (\emph{e.g.,}~image regions of interest and textual explanations) to support the answering process.
This explanatory mechanism enables VQA systems to reason for multimodal understanding.

\begin{figure}[t]
    \centering
    \includegraphics[trim= 10pt 92pt 36pt 92pt, clip=True, width=1\linewidth]{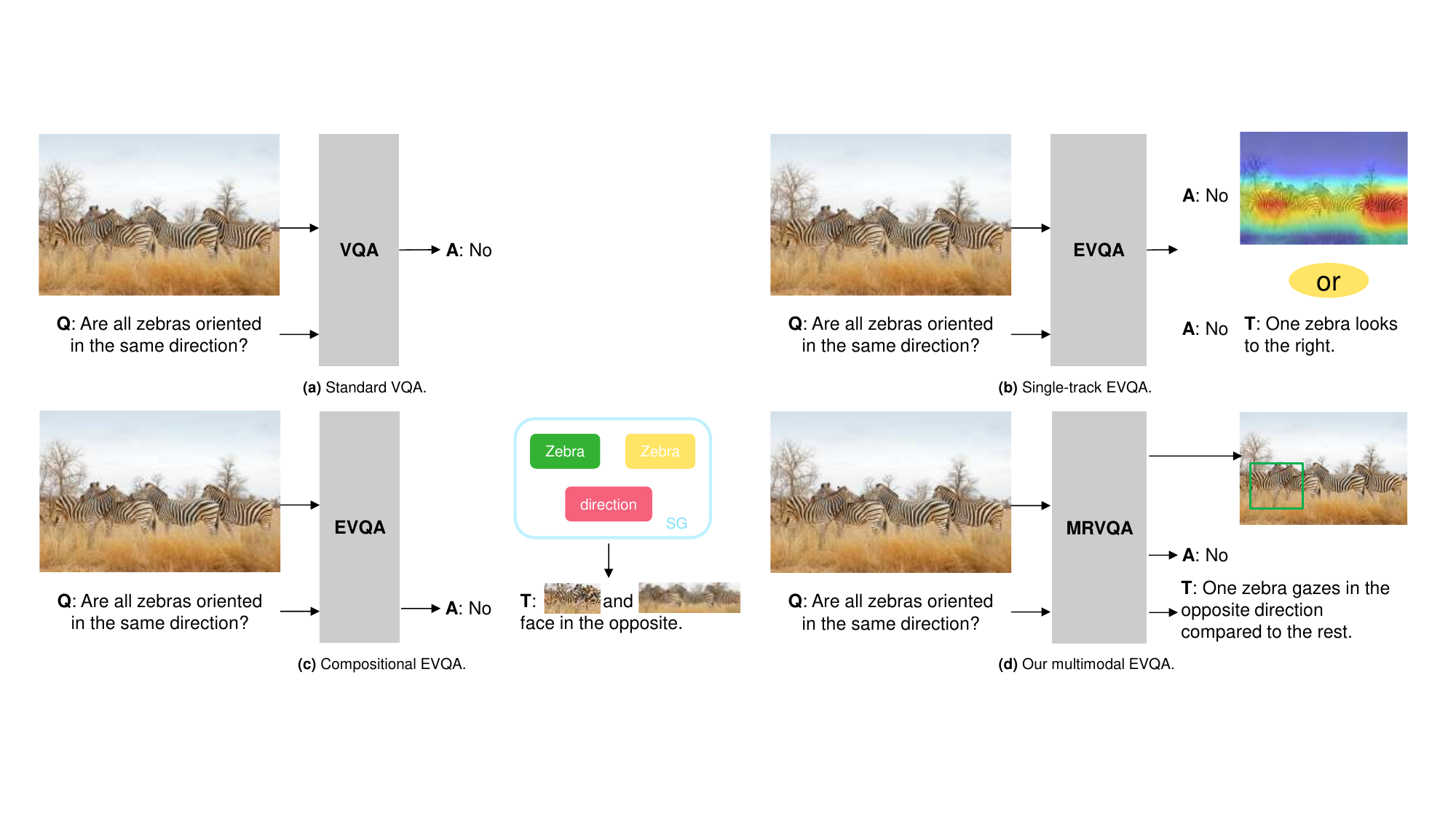}
    \caption{Illustration of the differences between standard VQA, single-track EVQA, compositional EVQA and our multimodal EVQA.
    Our MRVQA answers the question and simultaneously predicts textual and visual rationales to support the answer.}
    \label{fig:illustration}
    \vspace{-5mm}
\end{figure}

Significant progress has been made in the field of ``explainable artificial intelligence" (XAI) \cite{samek2019explainableAI}, uncovering the inner workings of AI algorithms and providing insights into how they arrive at their decisions.
However, there remains a largely unexplored area in developing human-centric explanations for interactive vision-language tasks, such as VQA.
Most of prevailing methods adopt single-track models to improve the explainability: either visual hints (termed ``V-Exp") or textual explanations (termed ``T-Exp"), as illustrated in Fig.~\ref{fig:illustration} (b).
V-Exp methods \cite{das2017vqahat, park2018multimodalexplanations} predict answers and further extract the regions of interest by analyzing the heat maps or attention maps in the inner neural network layers. 
However, the provided information is usually coarse and inaccurate while they are restricted to a region level, which is not human-friendly for understanding the decision-making process.
T-Exp methods \cite{li2018vqae, wu2019faithfulexplanation, zhang2021dmrfnet} aim to describe the visual content of the image that helps explaining the answer in natural language.
The explanation partially reproduces the reasoning process behind answering.
However, it still lacks a rich comprehension of vision-language inputs when we aim to directly and visually identify the specific target (\emph{i.e.,}~the zebra in the green bounding box from Fig.~\ref{fig:illustration} (d)). 
Another drawback arises from the inconvenience of textual explanations in complex scenes (\emph{e.g.,}~multiple zebras lying on the ground or the presence of horses), leading to longer and more detailed descriptions of the relevant objects.
Based on the above concerns in mind, we observe that the current single-track EVQA methods cannot satisfy the credibility of an interactive vision-language system.

In fact, prior to this paper, a few works \cite{chen2022rex, xue2023variationaliccv2023} have made significant strides towards EVQA with compositional reasoning.
However, these works essentially rely on various relationships among objects with external information (\emph{e.g.,}~scene graphs \cite{qian2022scenetmm}), resulting in a lack of visual recognition capabilities (as shown in Fig.~\ref{fig:illustration} (c)).
Specifically, the textual explanations are either chosen from multiple choices or generated using functional programs from a predefined candidate pool.
The shortcuts within the external information allow the model to directly utilize these relationships (\emph{e.g.,}~spatial relation and object category), which can harm the model's general performance.
In addition, the external information itself is expensive to collect.
To our knowledge, few previous studies have explored free-form textual explanations alongside precise visual predictions.

Moreover, the assessment of generated explanation quality remains limited to basic text similarity measures.
The existing methods \cite{li2018vqae, park2018multimodalexplanations, chen2022rex} for EVQA employ widely-used metrics for text generation evaluation, such as BLEU \cite{papineni2002bleu}, ROUGE \cite{lin2004rouge}, and CIDEr \cite{vedantam2015cider}.
Normally, these metrics quantitatively compare the generated text to reference texts from various textual aspects (\emph{e.g.,}~n-gram overlap, semantic similarity, word order, and structure), while ignoring the role of visual content in cross-modality tasks.
However, an ideal evaluation of multimodal explanations should consider both visual and linguistic perspectives.

Therefore, in this work, we attempt to address these problems with a multimodal EVQA model.
Specifically, we propose a new model called \textbf{MRVQA} (Multimodal Rationale for Visual Question Answering) that not only predicts accurate answers but also delivers both textual and visual rationales.
To highlight the differences between existing EVQA methods and our approach, an example is presented in Fig.~\ref{fig:illustration}.
Compared to single-track ones, our approach can simultaneously provide multimodal rationales, offering more reliable and user-friendly explanations.
For instance, the textual rationale ``One zebra gazes in the opposite direction" and the visual rationale, represented by the green bounding box, complement each other to support the answer prediction ``no".
Different from compositional ones, our approach generates free-form textual rationales and precise bounding boxes without relying on external information.
From a data perspective, we synthesize a multimodal rationale dataset by augmenting existing datasets to alleviate the annotation effort required by EVQA.
From a model perspective, we base our approach on a Transformer architecture \cite{NIPS2017_3f5ee243}, projecting features into the latent space of a large language model for textual predictions, and fusing features to extract bounding boxes by leveraging a pre-trained open-vocabulary detector.
Additionally, we propose a loss function designed to promote the effective cross-alignment and balance among multitask predictions.
From an evaluation perspective, a new metric is introduced, which incorporates both visual and textual views to assess generated rationales and advance EVQA research.

The \textbf{main contributions} of this paper are summarized as follows:
\begin{itemize}
   \setlength\itemsep{0.1em}
    \item We introduce \textbf{MRVQA}, a new EVQA model that facilitates the system to provide multimodal explanations for answer predictions through textual and visual rationale generation.
    Moreover, a new loss is introduced to balance multitask predictions and generate reasonable rationales.
    
    \item To assess the quality of generated rationales, we introduce a new metric called \textbf{vtS} for evaluating natural language from a visual perspective.
    
    \item We demonstrate the effectiveness of our model with extensive experiments under different settings. 
    The proposed model achieves superior performance over existing methods on three EVQA datasets.
\end{itemize}

\section{Related Work}
\label{sec: related work}
 
\subsection{Visual Question Answering}
\label{sec: related work vqa}
Visual Question Answering (VQA) has been defined as the task of providing an answer to a question about an image.
As this task takes as input both images and questions, VQA models address the problem through diverse approaches in different stages, including feature extraction \cite{anderson2018bottom}, alignment \cite{liu2019aligning1}, fusion \cite{ben2017mutan}, and prediction \cite{kafle2016answertypeprediction}.
With the rapid development of Transformers \cite{NIPS2017_3f5ee243}, both natural language processing and computer vision communities have witnessed remarkable advances in vision-language tasks \cite{qian2023locatetmm, li2024hrvqa}.
For instance, MCAN \cite{yu2019mcan} introduced a deep modular co-attention network by leveraging intra-modal self-attentions and cross-modal guided attentions, which enhanced the feature representation and alignment.
Some recent works \cite{lu2019vilbertpretraining, zhou2020unifiedpretraining} investigate leveraging the vision-language pre-training from large-scale data to enhance the efficacy of VQA models and enable complex reasoning, pioneering the burgeoning trend in the vision-language community.
These methods rely on extensive pre-training data and meticulous fine-tuning for downstream tasks, yet they still infer answers through a ``black-box" system, which diminishes the explainability of results. 

\subsection{Explanations for VQA}
\label{sec: related work evqa data}
Deep neural networks are typically ``black-box" systems because their internal mechanics and decision-making processes are not easily interpretable or understandable.
Some pioneering works have emerged that bring explanations to VQA.
Visualization is the first idea adopted to display the inherent details of models, aiding users in understanding the predictions.
For instance, VQA-HAT \cite{das2017vqahat} collected human attention maps of where humans choose to look so as to answer questions, visually providing coarse explanations for VQA.
Textual explanations utilize natural language to articulate the reasoning behind the system's predictions.
VQA-X \cite{park2018multimodalexplanations} combined textual explanations annotated by humans and visual pointing (similar to attention maps) to construct a multimodal reasoning dataset.
Subsequently, Li \textit{et al.} \cite{li2018vqae} further emphasized textual explanations by employing a semi-automatic construction method using caption annotations, creating a large-scale EVQA dataset (\emph{i.e.,}~VQA-E).
VCR \cite{zellers2019vcr} transferred the generation task to a simpler classification one by building a multiple choice movie-scene dataset.
GQA-REX \cite{chen2022rex} constructed a large-scale visually grounded textual explanation dataset by extending the previous GQA \cite{hudson2019gqa} dataset with an automatic scheme.
However, the displayed decision-making process heavily relies on the pre-defined functional programs with external information (\emph{e.g.,}~scene graphs and correlations), which reduces its applicability in a diverse open world.

\subsection{Generating Multimodal Explanations}
\label{sec: related work evqa model}
The increasing interest in understanding the reasoning behind VQA has led to the development of various methods across both textual and visual perspectives \cite{zellers2019vcr, chen2022rex}.
Park \textit{et al.} \cite{park2018multimodalexplanations} first adopted an LSTM \cite{hochreiter1997lstm} to generate textual explanations and subsequently displayed the internal visual focus through attention maps.
Li \textit{et al.} \cite{li2018vqae} employed two prediction heads to generate answers and textual explanations, respectively.
Wu \textit{et al.} \cite{wu2019faithfulexplanation} proposed to leverage the consistency between gradient-based visual explanations and answers to generate robust textual explanations.
These earlier works employ encoders to separately extract intra-modality features, which are then simply fused being passed to the final prediction stage. 
Marasović \textit{et al.} \cite{marasovic2020naturalrationalesemnlp} incorporated pretrained language models with object recognition, visual semantics, and scene graphs to generate textual explanations.
DMRFNet \cite{zhang2021dmrfnet} leveraged pre-trained semantic relation embeddings and adaptively fused spatial and semantic features through multi-graph reasoning and fusion layers.
REX \cite{chen2022rex} enhanced cross-modality alignment by considering the semantic similarity between words and image regions.
Subsequently, VCIN \cite{xue2023variationaliccv2023} extended relationships to include words, image regions, and explanation tokens through a gating transformer with variational causal inference.
However, these methods restrict the relations with pre-defined compositions (\emph{e.g.,}~semantic relations, scene graphs and functional programs).
In addition, visual explanations are usually generated through gradient-based segmentation proposals \cite{wu2019faithfulexplanation} or post-processed visual regions \cite{chen2022rex, xue2023variationaliccv2023}.
Different from the aforementioned methods, our proposed approach focuses on generating free-form textual explanations and object-level visual explanations.
\section{Method}
\label{sec: method}

\subsection{Problem Formulation}
\label{sec: formulation}
We first revisit the standard VQA setting.
Given an image $I$ and a natural language question $Q$, the desired output of a VQA model is an answer $A$ (\emph{i.e.,}~a single word or a short phrase) that relates the queried image.
To overcome the limitation in standard VQA, we expand the model's predictions beyond a single answer to include multimodal rationales, offering explanations from both visual and textual perspectives.
Regarding the textual part, we follow previous single-track EVQA methods \cite{li2018vqae, zhang2021dmrfnet} and employ a similar text description to explain the related context for answering the question, which we refer to as the textual rationale $TR$.
In contrast to previous methods \cite{chandrasekaran2018doexplanationshuman, park2018multimodalexplanations} that use region-level attention maps for visual explanations, we utilize precise bounding boxes of relevant objects as the visual rationale $VR$.
To this end, given $I$ and $Q$, the goal of our multimodal EVQA model is to predict multiple results: (1) an answer $A$; (2) a textual rationale $TR$; and (3) a visual rationale $VR$.
It is simply formulated as follows,
\begin{equation}
    \{A, TR, VR\} = P({I}, {Q}; \theta),
\end{equation}
where $\theta$ represents the parameters of an EVQA model.
It is particularly valuable for users who wish to explore more details in an interactive scenario.
Moreover, these rationales enhance the model's explainability and credibility.

\subsection{Multimodal EVQA Dataset Synthesis}
\label{sec: data synthesis}
Unfortunately, we have observed that the available EVQA datasets \cite{das2017vqahat, li2018vqae, park2018multimodalexplanations, zellers2019vcr, chen2022rex} lack either precise visual evidences or related textual descriptions.
To avoid the labor-intensive process of annotating explanation-aware VQA examples, we propose a semi-automatic approach to synthesize examples by augmenting an existing EVQA dataset.
We select VQA-E \cite{li2018vqae} for its comprehensive set of examples that include detailed textual explanations and, importantly, for its integration with COCO \cite{lin2014COCO} that provides precise bounding-box annotations for object detection.

Specifically, we first extract all nouns from the question-answer-text triplets in the VQA-E dataset.
These nouns usually identify the objects relevant to the explanation of answer predictions.
However, analyzing all nouns is impractical due to the presence of over 10K categories.
To address this, we conduct a frequency analysis of the nouns, selecting those that appear more than 20 times.
This step yields 1,247 distinct nouns that can be matched with the visual labels.
Unfortunately, COCO only includes annotations for 80 object categories.
To address this, we implement a strategy that classifies candidate nouns into broader categories aligned with the existing COCO annotations.
For instance, nouns like ``man", ``woman", ``people", and ``girl" are categorized under the ``Person" category in COCO. 
However, not all bounding boxes within a given category are relevant for explanations of a specific image-question pair.
For example, although two sheep are present in an image, only one serves as the correct visual rationale for a question concerning the ``right sheep".
Additionally, noun-based object extraction may miss key contextual elements. For example, in answering ``What is the man doing?” with ground truth ``surfing” and ``a man on the wave”, only ``man” yields a bounding box, while the full rationale should include both the man and the surfboard.
To ensure the accuracy and relevance of visual rationales, two annotators, following detailed guidelines, manually refined the bounding boxes over two weeks by removing irrelevant objects and adding those essential to the explanation.
This process results in the synthesis of a dataset (VQA-E-Syn).
Overall, it comprises 33,726 question-answer pairs derived from 20,367 images, adhering to the original split in VQA-E.
Each pair is accompanied by a textual rationale and an average of 2.8 bounding boxes for visual rationales.

\begin{figure*}[t]
    \centering
    \includegraphics[trim= 10pt 168pt 10pt 162pt, clip=True, width=1\linewidth]{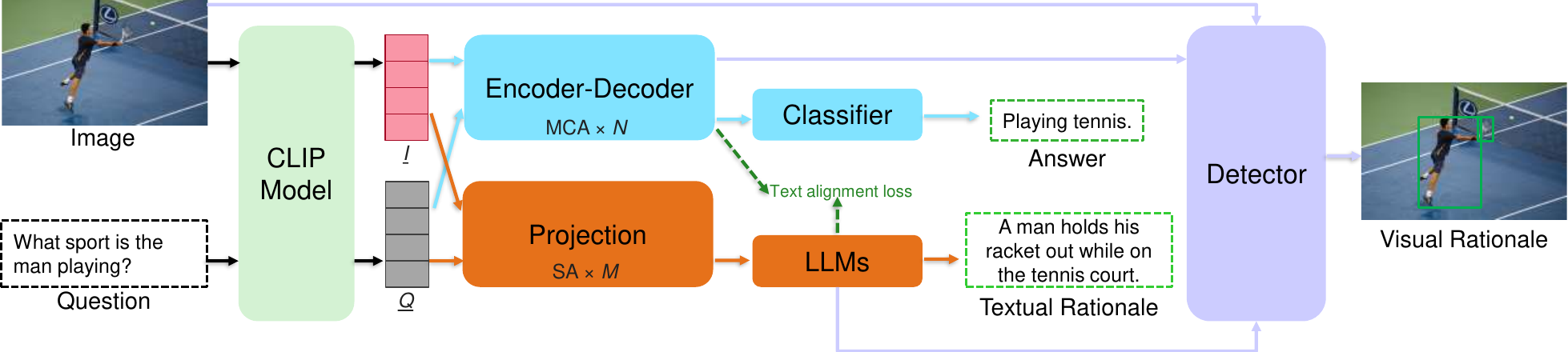}
    \caption{The framework of the proposed MRVQA model.
    The notations $\underline{I}$ and $\underline{Q}$ represent the pre-trained CLIP \cite{radford2021clip} features for the image and the question, respectively.
    The proposed text alignment loss helps to improve the coherence of text predictions.
    }
    \label{fig:crvqa}
    \vspace{-5mm}
\end{figure*}

\subsection{Multimodal EVQA Network}
\label{network}
The overall framework of the proposed MRVQA model is shown in Fig.~\ref{fig:crvqa}.
Here, the core components of our approach are introduced as follows.

\textbf{Input Representation.}
Different from previous methods that use separate visual and linguistic feature extractors, we utilize a pre-trained vision-language model (\emph{i.e.,}~CLIP \cite{radford2021clip}) to simultaneously represent both the input image and question. 
The CLIP model, pre-trained on large-scale image-caption pairs and benefited from transferable representations, outperforms intra-encoders trained solely on in-domain data.
With the pre-trained CLIP model, we generate robust visual and linguistic features for our EVQA task.

\textbf{Encoder-Decoder.}
To effectively understand the queried focus from the visual and linguistic features, it is essential to exploit cross-modality fusion for VQA tasks.
To enhance representations without introducing a new backbone or altering the model’s structure, we build on the popular VQA baseline MCAN \cite{yu2019mcan}, which cascades several Modular Co-Attention (MCA) layers.
The MCA layer comprises self-attention units for intra-modal interaction and guided-attention units for cross-modal interaction.

\textbf{Projection Module.}
To alleviate the mismatch between the pre-trained CLIP  and our textual rationale generator (Large Language Models, LLMs described in the following part), we employ a Transformer-based \cite{NIPS2017_3f5ee243} projection module, facilitating multimodal adaptation and integration.
The self-attention mechanism enables the learning of global dependencies between tokens, even within the context of long sentences.
Inspired by Darcet et al. \cite{darcet2024visionregisters}, we introduce two additional tokens as learnable constants for the CLIP features $\underline{I}$ and $\underline{Q}$, respectively.
These constant tokens act as registers, allowing the model to store and retrieve meaningful information efficiently.
For the entire projection module, we utilize eight self-attention layers to process the dual-stream visual and linguistic features.
The projected features $I_{proj}$ and $Q_{proj}$ are calculated as follows,
\begin{equation}
    I_{proj} = SA(Cat(\underline{I}, C)),  Q_{proj} = SA(Cat(\underline{Q}, C)),
\end{equation}
\begin{equation}
    SA(q, K, V) = Softmax(\frac{qK}{\sqrt{d}})V,
\end{equation}
where $C$ represents the learnable constants, q, K, V and d denote query, Key, Value, and dimensionality of values, respectively.
The projected features $I_{proj}$ and $Q_{proj}$ are fused by concatenation as $f_{proj}$ before fed into the LLMs.

\textbf{Predictors.}
For the answer prediction, we approach it as a classification problem and utilize a linear fusion module, followed by a sigmoid function, to process the cross-attended features from the decoder.
For textual rationales, in contrast to earlier methods that rely on simple LSTMs \cite{hochreiter1997lstm}, we utilize advanced LLMs to generate more robust and relevant textual rationales.
However, given the complexity of handling multiple tasks in EVQA, the LLMs integrated into the model should facilitate computationally efficient fine-tuning for multimodal learning.
To address this, we leverage the auto-regressive large language model GPT-2 \cite{radford2019gpt2}.
For each token in the sequence, GPT-2, conditioned on the projected features $f_{proj}$, generates probability distributions over the entire vocabulary and predicts the next token in the sequence.
For visual rationales, we generate precise object-level bounding boxes using Grounding-DINO \cite{liu2023grounding-dino}, an open-vocabulary object detector based on text.
In contrast to the original work, our approach inputs the detector with the image $I$, cross-attended features $f_{cross}$, and LLM-processed features $f_{llm}$.
In addition, we set all detected bounding boxes as a single category.
This setting addresses the challenge of aligning objects with the input question for discerning the category information from the image-question pair due to the lack of explicit context.

\textbf{Loss Functions.}
A hybrid loss function is employed to refine the proposed model, with different loss components tailored for each type of prediction in our EVQA task.
Consistent with previous methods \cite{li2018vqae}, we employ the widely-used binary cross-entropy loss
\cite{teney2018tips} as $L_{ans}$ to supervise answer predictions.
To train the model for generating textual rationales from the LLM-processed features $f_{llm}$, we utilize the simple yet effective cross-entropy loss as $L_{tr}$ in our case.
Given $M$ questions, $N$ candidate answers, $M$ textual rationales, $L_{ans}$ and $L_{tr}$ are calculated as follows,
\begin{equation}
    L_{ans} = -\sum_i^M\sum_j^Ns_{ij}\log(\hat{s}_{ij}) - (1 - s_{ij})\log(1 - \hat{s}_{ij}),
\end{equation}
\begin{equation}
    L_{tr} = -\sum_i^M\sum_t^l\log p_\theta(w_t^i|f_{llm}^i, w_1^i, ..., w_{t-1}^i),
\end{equation}
where $l$ and $\theta$ represent the maximum length of tokens and the model's trainable parameters, respectively.
The sequenced tokens are denoted as $w^i = w_1^i, ..., w_l^i$.
For visual rationale generation, we directly adopt the existing loss function from Grounding-DINO \cite{liu2023grounding-dino} as $L_{vr}$, including L1 loss, GIoU loss, and contrastive loss.
To train the entire model using multi-task learning, we take a linear combination of the different loss components.

Moreover, to improve the coherence between answers and their corresponding textual rationales, we introduce a new loss function termed \textbf{text alignment loss} (abbreviated as $L_{ta}$).
Specifically, the improvement of text coherence is achieved by the alignment between cross-attended features $f_{cross}$ and LLM-processed features $f_{llm}$.
We use the cosine similarity function to measure the feature similarity between these dual objectives from the textual side.
The text alignment loss $L_{ta}$ is obtained as follows,
\begin{equation}
    L_{ta} = 1 -\sum_i^M\sum_t^lcos(f_{llm}'^i, f_{cross}'^i,),
\end{equation}
where $f_{cross}'$ and $f_{llm}'$ represent the co-matched features in a common vector space through two MLP layers.
Before incorporating it into the final loss function, we use a trade-off parameter $\lambda$ to balance the basic and alignment components of the loss.
The overall loss $L$ of the model is defined as follows,
\begin{equation}
    L = L_{ans} + L_{tr} + L_{vr} + \lambda L_{ta},
\end{equation}

\subsection{Visual-Textual Similarity Score}
\label{sec: metric}
We utilize the powerful pre-trained text embedding model, GTE \cite{li2023textsimilarity}, to map the generated textual rationales for similarity comparison to the ground truths.
There are several reasons for selecting this model: (1) The GTE model employs contrastive learning across a diverse range of datasets, which allows it to generalize effectively to various data; (2) It avoids the reliance on prompt formulation seen in LLMs; (3) It handles pairs of short and long texts more effectively without requiring word-level alignments, which is ideal for comparing our generated free-form texts; (4) It outperforms previous embedding methods in comprehensive benchmarks \cite{li2023textsimilarity}.
Specifically, we use the cosine function to assess textual similarity.
To handle the negative values produced by the cosine function, we map the cosine similarity scores to a range from 0 to 1.
The textual similarity ($TS$) is calculated as follows,
\begin{equation}
    TS = \frac{1 + \cos{({\text{GTE}(T_{pred}), \text{GTE}(GT)})}}2,
\end{equation}
where $T_{pred}$ and $GT$ represent the generated textual rationale and the ground truth, respectively.
We assess the predicted bounding boxes by calculating the Average Precision \cite{everingham2010ap} (AP) in comparison to ground truths.

To this end, we introduce a new metric \textbf{vtS} (\textbf{v}isual-\textbf{t}extual \textbf{S}imilarity score), which combines visual evaluation and textual similarity to provide a balanced assessment of the generated rationale quality.
The metric \textbf{vtS} is defined as follows,
\begin{equation}
    vtS = \frac{2 \times TS \times AP}{TS + AP},
\end{equation}
\section{Experiments}
\label{sec: experiments}
\subsection{Experiment Setup}
\label{sec: setup}
\textbf{Datasets.}
We conducted primary experiments on the synthesized multimodal EVQA dataset (VQA-E-Syn) introduced in Section~\ref{sec: data synthesis}.
In addition, we evaluated the model on two existing EVQA datasets: VQA-X \cite{park2018multimodalexplanations} and GQA-REX \cite{chen2022rex}, to access the its adaptability and generalization capability under different settings.

\textbf{Training and Inference Settings.}
Under full supervision, we refined the model in an end-to-end manner, referred to as MRVQA-E.
To adapt to various practical scenarios, we simplified the MRVQA model to predict the required results under different conditions.
For example, we developed a variant, MRVQA-C (combined), by removing the visual generator component, allowing us to test the model's ability to generalize in the absence of visual rationale supervision.
This variant utilized the generated textual rationales and input images to guide the detector in predicting bounding boxes during inference stage.
In addition, we further simplified the model to focus exclusively on answer prediction or textual rationale prediction, developing variants called MRVQA-A and MRVQA-TR, respectively.

\begin{table*}\footnotesize
  \caption{Comparison results on VQA-E-Syn regarding textual rationale generation.
  All numbers are reported in percentage (\%). The best results are \textbf{bold} while the second best are \underline{underlined}.}
  \centering
  \setlength\tabcolsep{5pt}
  \begin{tabular}{p{2.5cm}ccccccccc}
  \toprule
  Method & Image Feature & Question Feature & Text Predictor & BLEU-4 & METEOR &  ROUGE & CIDEr & SPICE & vtS\\ 
 \midrule
  PJ-X \cite{park2018multimodalexplanations}  & CNN & LSTM & LSTM & 8.78 & 16.94 & 35.65 & 89.31 & 15.32 & 49.19 \\ 
  VQA-E \cite{li2018vqae}  & CNN & GRU & LSTM & 8.93 & 17.02 & 35.96 & 90.84 & 16.83 & 51.88\\ 
  FME \cite{wu2019faithfulexplanation}  & CNN & GRU & LSTM & 12.81 & 21.26 & 37.69 & 89.16 & 18.77 & 53.82\\ 
  CCM \cite{patro2020robustwacv}   & CNN & LSTM & LSTM & 8.85 & 17.86 & 38.42 & 92.46 & 18.35 & 54.44\\ 
  DMRFNet \cite{zhang2021dmrfnet}  & CNN & GRU & GRU & 13.34 & 19.44 & 40.76 & 95.68 & 20.41 & 56.06\\ 
  REX \cite{chen2022rex}  & VilBert & VilBert & LSTM & 14.71 & 21.35 & 41.86 & 100.75 & 21.08 & 56.38\\ 
  OFA \cite{wang2022ofa}  & CNN & BPE & Seq2Seq & \underline{16.38} & 22.09 & 42.74 & 103.25 & \underline{22.65} & 56.58\\ 
  VCIN \cite{xue2023variationaliccv2023} & V-Bert & V-Bert & Trans & 15.77 & 22.38 & 43.10 & 104.63 & 22.07 & 57.89\\
  MRVQA-TR (ours)  & CLIP & CLIP & GPT2 & 15.08 & 21.77 & 42.04 & 102.86 & 21.22& 57.30\\ 
  MRVQA-C (ours)  & CLIP & CLIP & GPT2 & 15.84 & \underline{22.41} & \underline{43.57} & \underline{105.31} & 22.19 & \underline{58.19}\\ 
  MRVQA-E (ours) & CLIP & CLIP & GPT2 & \textbf{16.97} & \textbf{23.02} & \textbf{44.28} & \textbf{107.04} & \textbf{23.68} & \textbf{59.16}\\ 
  \bottomrule
  \end{tabular}
  \label{tab: vqar comparison with SOTA for tr}
  \vspace{-5mm}
\end{table*}

\textbf{Implementation Details.}
We utilized the pre-trained CLIP model \cite{radford2021clip} with ViT-base \cite{dosovitskiy2020vit} and Transformer \cite{NIPS2017_3f5ee243} for input representation.
The AdamW \cite{kingma2014adamw} optimizer was used with a learning rate of $2 \times 10^{-5}$.
All the models were trained for 50 epochs with a batch size of 64 using two A40 GPUs, within a consistent PyTorch environment.

\textbf{Evaluation Metrics.}
For answer evaluation, we maintained consistency with prior works by using accuracy as the metric.
To evaluate generated textual rationales, we employed five mainstream metrics, namely BLEU-4 \cite{papineni2002bleu}, METEOR \cite{banerjee2005meteor}, ROUGE \cite{lin2004rouge}, CIDEr \cite{vedantam2015cider} and SPICE \cite{anderson2016spice} (B, M, R, C, S for their abbreviations, respectively).
Additionally, the newly introduced metric \textbf{vtS} was used to capture the cross-modality similarity by leveraging the visual response.

\subsection{Quantitative Results and Comparison}
\label{quantitative comparison}

\subsubsection{Results on the synthesized VQA-E-Syn}
\label{res: vqar}
Table~\ref{tab: vqar comparison with SOTA for tr} presents a comparison with state-of-the-art EVQA methods trained and evaluated on the synthesized VQA-E-Syn dataset.
Our proposed MRVQA-E model significantly enhanced the generation of textual rationales for multimodal EVQA, outperforming the previous methods.
For example, our approach achieved a 2.41\% improvement in CIDEr compared to VCIN \cite{xue2023variationaliccv2023}, highlighting its superior fluency and coherence.
To ensure a fair comparison with previous methods lacking visual prediction capabilities, we adopted a combined approach similar to our MRVQA-C variant.
The results demonstrate that our approach achieved a 1.27\% improvement over VCIN in vtS.
Additionally, we also evaluated our MRVQA-TR and MRVQA-C models.
Both variants demonstrated slightly superior performance compared to the previous methods.
More importantly, the comparison between the end-to-end model and its variants highlights that incorporating visual rationale generation improves linguistic prediction performance.

\begin{table}\footnotesize
  \caption{Comparison results on VQA-E-Syn regarding answer accuracy.
  OA and AA represent overall and average accuracy, respectively.
  }
  \centering
  \setlength\tabcolsep{5pt}
  \begin{tabular}{p{2.1cm}ccccc}
  \toprule
  Method & Number & Yes/No & Other &  OA & AA\\
 \midrule
  MUTAN \cite{ben2017mutan}  & 49.23 & 65.52 & 60.27 & 61.07 & 58.34\\
  BUTD \cite{anderson2018bottom}  & 52.68 & 64.07 & 65.72 & 64.13 & 60.83 \\
  MCAN \cite{yu2019mcan} & 59.19 & 71.70 & 72.37 & 71.09 & 67.76\\
  LXMERT \cite{tan2019lxmert}  & 59.93 & 72.34 & 73.04 & 71.76 & 68.44\\
  UNITER \cite{chen2020uniter}  & 61.58 & 73.81 & 72.66 & 72.14 & 69.35\\
  BLIP \cite{li2022blip} & \underline{63.68} & 72.43 & 74.83 & 73.16 & 70.31\\
  CMQEF \cite{li2024ifcontextvqa}  & 60.02 & 71.63 & 72.20 & 71.03 & 67.95\\
  \hline
  VQA-E \cite{li2018vqae}  & 54.26 & 63.98 & 66.45 & 64.67 & 61.57\\
  DMRFNet \cite{zhang2021dmrfnet}  & 59.08 & 72.28 & 73.84 & 72.15 & 68.40\\
  REX \cite{chen2022rex}  & 59.73 & 72.91 & 74.18 & 72.84 & 68.94\\
  OFA-base \cite{wang2022ofa}  & 60.48 & 73.26 & 75.37 & 73.49 & 69.70\\
  VCIN \cite{xue2023variationaliccv2023} & 61.59 & 73.47 & 73.10 & 72.43 & 69.39\\
  \hline
  MRVQA-A (ours)  & 57.02 & 72.80 & 73.41 & 71.89 & 67.75\\
  MRVQA-C (ours) & 63.20 & \underline{76.03} & \underline{75.72} & \underline{74.81} & \underline{71.65} \\
  MRVQA-E (ours) & \textbf{65.11} & \textbf{76.40} & \textbf{75.87} & \textbf{75.01} & \textbf{72.46} \\
  \bottomrule
  \end{tabular}
  \label{tab: vqar comparison with SOTA for answer}
  \vspace{-3mm}
\end{table}

Table~\ref{tab: vqar comparison with SOTA for answer} shows the results in terms of answer accuracy.
The methods listed in the top section are trained solely for answer prediction, whereas those in the middle section have the capability to generate both answers and textual explanations.
OFA \cite{wang2022ofa}, as a powerful unified sequence-to-sequence framework, facilitates generating multiple predictions through separate fine-tuning processes on task-specific datasets (\emph{e.g.,}~VQA and image captioning).
To adapt OFA for our task, generating texts within a unified inference pass, we performed a single fine-tuning of the OFA-base model on the combined samples from VQA-E-Syn.
Methods lacking answer prediction details or accuracy results \cite{park2018multimodalexplanations, wu2019faithfulexplanation, patro2020robustwacv} are excluded from this table.
Our MRVQA-E model achieved an overall accuracy of 75.01\% and an average accuracy of 72.46\% on VQA-E-Syn, surpassing the previous methods.
Notably, the model showed significant improvements in the \textit{Number} and \textit{Yes/No} question categories, with increases of 1.43\% and 2.59\%, respectively.
Similarly, comparing MRVQA-E with its variants (last three rows) reveals that incorporating additional rationale supervision enhanced the performance of the answering process.

\subsubsection{Results on VQA-X}
\label{res: vqax}
Table~\ref{tab: vqax comparison with SOTA for tr} reports the results compared to previous EVQA methods on VQA-X \cite{park2018multimodalexplanations}.
As VQA-X only supports linguistic examples, to ensure a fair comparison, we conducted experiments with MRVQA-C for only answer and textual rationale generation.
Our approach achieved a 78.8\% overall accuracy, which is higher than the previous state-of-the-art VCIN by 1.1\%.
Also, the model showed significant improvements in BLEU-4 and CIDEr metrics, with increases of 0.7\% and 4.7\%, respectively, compared to VCIN.

\begin{table}\footnotesize
  \caption{Comparison results on VQA-X \cite{park2018multimodalexplanations} and GQA-REX \cite{chen2022rex} for answer and textual rationale evaluation.}
  \centering
  \setlength\tabcolsep{2pt}
  \begin{tabular}{p{1.3cm}ccccccc}
  \toprule
  Dataset &  Method & B & M &  R & C &  S & OA \\
  \midrule
  \multirow{7}{*}{VQA-X} & PJ-X \cite{park2018multimodalexplanations} & 19.5 & 18.2 & 43.7 & 71.3 & 15.1 & -\\
  & VQA-E \cite{li2018vqae}  & 20.7 & 18.6 & 44.2 & 75.6 & 15.6 & 70.2\\
  & FME \cite{wu2019faithfulexplanation}  & 24.4 & 19.5  & 47.4 & 88.8  & 17.9 & - \\
  & CCM \cite{patro2020robustwacv} & 21.1 & 19.7 & 44.9 & 73.9  & 16.2 & - \\
  & DMRFNet \cite{zhang2021dmrfnet} & 20.5 & 19.9 & 41.3 & 74.5 & 17.6 & 72.6 \\
  & VCIN \cite{xue2023variationaliccv2023} &  25.9 & 21.6 & 48.5 & 93.7 & \textbf{19.4} & 77.7 \\
  & MRVQA-C  & \textbf{26.6} & \textbf{22.0} & \textbf{48.9} & \textbf{98.4} & 19.2 & \textbf{78.8}\\
  
  \hline
  \multirow{5}{*}{GQA-REX} & VQA-E \cite{li2018vqae}  & 42.56 & 34.51 & 73.59 & 358.20 & 40.39 & 57.24\\
  &  FME \cite{wu2019faithfulexplanation}  & 42.45 & 34.46  & 73.51 & 357.10  & 40.35 & 56.92 \\
  & REX \cite{chen2022rex} & 54.59 & 39.22 & 78.56 & 464.20 & 46.80 & 57.77 \\
  & VCIN \cite{xue2023variationaliccv2023} &  58.65 & 41.57 & 81.45 & 519.23 & 54.63 &  60.61\\
  & MRVQA-C  & \textbf{58.83} & \textbf{42.01} & \textbf{83.07} & \textbf{528.42} & \textbf{54.96} & \textbf{61.28}\\
  
  \bottomrule
  \end{tabular}
  \label{tab: vqax comparison with SOTA for tr}
  \vspace{-3mm}
\end{table}

\subsubsection{Results on GQA-REX}
\label{res: rex}
Table~\ref{tab: vqax comparison with SOTA for tr} also shows the results on GQA-REX \cite{chen2022rex}.
Note that the previous VCIN model predicts region-level patches, which relies on priors in visual encodings (\emph{e.g.,}~patch location).
Due to the single-category object detection used in our settings, we exclude this unfair comparison and only report the textual results.
Our approach achieved an overall answer accuracy of 61.28\%, representing a 0.67\% improvement over the previous state-of-the-art VCIN.
Meanwhile, our model maintained competitive performance even without external information, achieving 58.83\% and 83.07\% for BLEU-4 and ROUGE scores, respectively.

\subsection{Ablation Studies}
\label{sec: ablation}
\subsubsection{Ablation of the Projection Module}
For comparison, we conducted an ablation study by replacing the projection module in MRVQA-E with a linear layer, as used in prior vision-language methods.
The results in Table~\ref{tab: linear or transformer} show that our Transformer-based projection module significantly enhanced the performance of textual rationale generation compared to the linear layer across various metrics.
For example, the incorporation of our module resulted in a 5.46\% improvement in CIDEr score.
Additionally, the enhancement in linguistic predictions contributed to a 2.02\% increase in the multimodal rationale evaluation metric vtS, highlighting the efficacy of the module.

\begin{table}\footnotesize
  \caption{Ablation results of using different projection modules for multimodal rationale generation on VQA-E-Syn.
  }
  \centering
  \setlength\tabcolsep{4pt}
  \begin{tabular}{p{2.1cm}cccccc}
  \toprule
  Module & B & M &  R & C &  S &  vtS\\
 \midrule
  Linear \cite{li2023LLM2023} & 15.07 & 22.13 & 42.74 & 101.58 & 22.06 & 57.14\\ 
  Ours-Transformer  & \textbf{16.97} & \textbf{23.02} & \textbf{44.28} & \textbf{107.04} & \textbf{23.68} & \textbf{59.16}\\ 
  \bottomrule
  \end{tabular}
  \label{tab: linear or transformer}
  \vspace{-3mm}
\end{table}

\subsubsection{Ablation of Unified or Separate Text Predictor}
Some approaches \cite{li2023blip2} use a unified strategy for all text-related downstream tasks, which often results in longer and more complex expressions.
To compare it with our approach, we adapt the unified strategy by integrating answer prediction into the text generation, using samples in \{answer, textual rationale\} format (\emph{e.g.,}~``the boy because only the boy can catch the frisbee").
Table~\ref{tab: vqar comparison with SOTA for text} reports the performance of different text prediction strategies.
While the performance of textual rationale generation showed a slight change, the overall accuracy significantly dropped by 4.39\%, indicating a decrease in the robustness of an answering model.
Thus, we employed dual-head predictors for linguistic outputs to address this issue in the EVQA task.

\begin{table}\footnotesize
  \caption{Ablation results of different strategies for text predictions on VQA-E-Syn.
  }
  \centering
  \setlength\tabcolsep{4pt}
  \begin{tabular}{p{2.7cm}ccccc}
  \toprule
  Strategy & Data & B & S & OA & AA\\
 \midrule
  Generator & \{A + TR\} & 15.16 & 20.98 & 70.42 & 68.39 \\
  Classifier + Generator & \{A, TR\} & \textbf{15.84} & \textbf{22.19} & \textbf{74.81} & \textbf{71.65} \\
  \bottomrule
  \end{tabular}
  \label{tab: vqar comparison with SOTA for text}
  \vspace{-3mm}
\end{table}

\subsubsection{Ablation of the Text Generator}
To evaluate LLMs' effectiveness compared to commonly used text generation models, such as LSTMs and Transformer, we replaced this component in MRVQA-E with these alternative modules.
The results are shown in Table~\ref{tab: frozen or finetuning}.
Comparing the first three rows, GPT-2 significantly outperformed standard LSTMs and yielded slightly better results than Transformer, even with frozen parameters.
Subsequently, we allowed the parameters of GPT-2 to be trainable to boost generation performance.
While this required additional effort to train the LLMs, it enhanced the quality of the generated textual rationales.
For instance, it resulted in a 13.78\% improvement in the CIDEr score.
This advancement in linguistic understanding also positively impacted visual predictions, as evidenced by a 5.82\% increase in the vtS score.
We further replaced the text generator with the more advanced GPT-3 model \cite{brown2020gpt3}, known for its superior performance in complex reasoning and contextual understanding.
However, GPT-3's size (175 billion parameters compared to GPT-2's 1.5 billion) poses significant challenges, and moreover, it is not available for fine-tuning.
We utilized GPT-3 with frozen parameters for generating textual rationales.
The results indicate that using more advanced LLMs does not significantly impact the performance of text generation for our EVQA task.

\begin{table}\footnotesize
  \caption{Ablation results of different text generators under either freezing or fine-tuning settings on VQA-E-Syn.
  }
  \centering
  \setlength\tabcolsep{2pt}
  \begin{tabular}{p{2cm}ccccccc}
  \toprule
  Generator & Mode & B & M &  R & C &  S & vtS\\
 \midrule
  LSTM \cite{hochreiter1997lstm} & - & 12.65 & 18.39 & 36.77 & 88.82 & 18.45 & 53.14\\
  Transformer \cite{NIPS2017_3f5ee243} & - & 12.89 & 18.91 & 37.03 & 89.14 & 18.86 & 53.17\\
  GPT-2 \cite{radford2019gpt2} & freezing & 13.27 & 20.08 & 38.33 & 93.26 & 19.25 & 53.34\\
  GPT-3 \cite{brown2020gpt3} & freezing & 15.06 & 21.27 & 39.74 & 98.90 & 20.57 & 56.25\\
  MRVQA-E & fine-tuning & \textbf{16.97} & \textbf{23.02} & \textbf{44.28} & \textbf{107.04} & \textbf{23.68} & \textbf{59.16}\\
  \bottomrule
  \end{tabular}
  \label{tab: frozen or finetuning}
  \vspace{-3mm}
\end{table}



\subsubsection{Ablation of Loss Functions}
To further evaluate the proposed loss function, we conducted ablation experiments on VQA-E-Syn to verify its effectiveness.
The results are reported in Table~\ref{tab: vqar loss function}.
Comparing the first two rows, we can see that incorporating visual supervision significantly improved overall performance across all types of predictions, which is consistent with our earlier observations in Table~\ref{tab: vqar comparison with SOTA for tr} and Table~\ref{tab: vqar comparison with SOTA for answer}.
Comparing the last four rows, we observed a notable enhancement in performance due to the inclusion of our proposed text alignment loss.
For example, all configurations utilizing $Loss_{TA}$ showed at least a 1\% improvement in SPICE scores over the others.
In addition, we performed a series of experiments to determine the optimal value for the trade-off parameter $\lambda$.
The results from the last three rows indicate that setting $\lambda=0.5$ yielded the best performance.

\begin{table}\footnotesize
  \caption{Ablation results of loss functions on VQA-E-Syn.}
  \centering
  \setlength\tabcolsep{3pt}
  \begin{tabular}{cccccccc}
  \toprule
  $Loss_A$ & $Loss_{TR}$ & $Loss_{VR}$ & $Loss_{TA}$ & B & S & vtS & OA\\
 \midrule
 \checkmark & \checkmark & &  & 15.84 & 22.19 & - & 71.65\\
  \checkmark & \checkmark & \checkmark &  & 15.13 & 22.34 & 59.01 & 72.28\\
  \checkmark & \checkmark & \checkmark  & $\lambda=0.1$ & 16.49 & 23.51 & 59.10 & 74.76\\
  \checkmark & \checkmark  & \checkmark  & $\lambda=0.5$ & \textbf{16.97} & \textbf{23.68} & \textbf{59.16} & \textbf{75.01}\\ 
  \checkmark & \checkmark  & \checkmark  & $\lambda=1$ & 16.25 & 23.40 & 59.07 & 74.59\\

  \bottomrule
  \end{tabular}
  \label{tab: vqar loss function}
  \vspace{-5mm}
\end{table}

\begin{figure}[t]
    \centering
    \includegraphics[trim= 182pt 3pt 240pt 3pt, clip=True, width=1\linewidth]{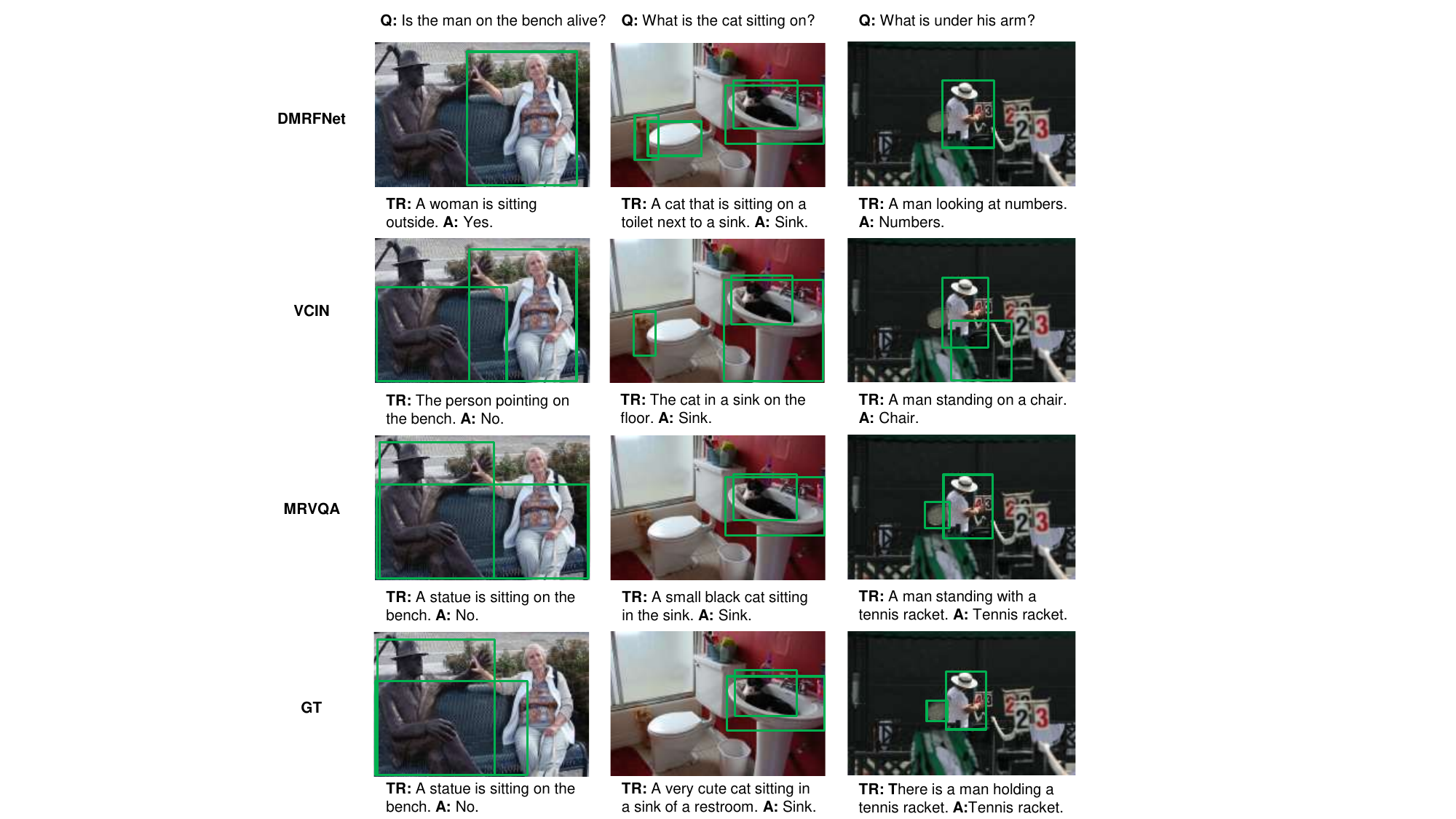}
    \caption{Qualitative comparison of results. The bounding boxes in \textcolor{green}{green} display the relevant visual rationales.}
    \label{fig: comparison}
    \vspace{-5mm}
\end{figure}

\subsection{Qualitative Analysis}
\label{qualitative}
Figure~\ref{fig: comparison} displays three examples from the synthesized multimodal dataset VQA-E-Syn.
Compared to the previous methods such as DMRFNet \cite{zhang2021dmrfnet} and VCIN \cite{xue2023variationaliccv2023}, our proposed approach demonstrates superior performance in both answering and rationale support.
For example, in the first column, both DMRFNet and VCIN struggled to comprehend the query's intent, focusing on the object ``woman," although VCIN ultimately predicted the correct answer ``No".
In contrast, MRVQA effectively performed cross-modal understanding, accurately identifying the target ``man" referring to the statue and correctly predicting the answer.
The medium column illustrates our model's ability to capture relevant objects in both visual and textual rationales.
It successfully identified the action ``sitting on" and located the target, while the other methods were misled by ``the other cat" or ``the toilet".
In the last example, despite the lack of category-level information, MRVQA accurately predicted the answer and delivered consistent rationales, as evidenced by the references to ``the man" and ``the tennis racket".
\section{Conclusion}
\label{sec: conclusion}
In this paper, we propose a multimodal EVQA method that predicts not only answers but also textual and visual rationales to support the answers.
To minimize the annotation effort, we synthesize a dataset by augmenting existing datasets through a semi-automatic matching engine.
To solve the multimodal EVQA problem, we introduce MRVQA that generates textual rationales with large language models and visual rationales via an pre-trained detector.
A loss function is introduced to ensure the consistency between the answer and textual rationale generation.
Furthermore, we propose a metric for evaluating the quality of generated rationales from a visual perspective.
The extensive experiments across three EVQA datasets demonstrate that MRVQA significantly outperforms existing state-of-the-art EVQA methods.
We believe this work will inspire further research in the EVQA field.

\noindent
\textbf{Acknowledgement} Kun Li has been supported by China Scholarship Council. Michael Ying Yang has been supported by EU HORIZON-CL42023-HUMAN-01-CNECT XTREME (grant no.101136006).
\newpage
{
    \small
    \bibliographystyle{ieeenat_fullname}
    \bibliography{main}
}


\end{document}